# Semi-supervised Semantic Segmentation of Prostate and Organs-at-Risk on 3D Pelvic CT Images

Zhuangzhuang Zhang, Tianyu Zhao, Hiram Gay, Weixiong Zhang*, and Baozhou Sun*

**Abstract**

Automated segmentation can assist radiotherapy treatment planning by saving manual contouring efforts and reducing intra-observer and inter-observer variations. The recent development of deep learning approaches has revoluted medical data processing, including semantic segmentation, by dramatically improving performance. However, training effective deep learning models usually require a large amount of high-quality labeled data, which are often costly to collect. We developed a novel semi-supervised adversarial deep learning approach for 3D pelvic CT image semantic segmentation. Unlike supervised deep learning methods, the new approach can utilize both annotated and un-annotated data for training. It generates un-annotated synthetic data by a data augmentation scheme using generative adversarial networks (GANs). We applied the new approach to segmenting multiple organs in male pelvic CT images, where CT images without annotations and GAN-synthesized un-annotated images were used in semi-supervised learning. Experimental results, evaluated by three metrics (Dice similarity coefficient, average Hausdorff distance, and average surface Hausdorff distance), showed that the new method achieved either comparable performance with substantially fewer annotated images or better performance with the same amount of annotated data, outperforming the existing state-of-the-art methods.

## 1. Introduction

Accurate delineation of targets and organs-at-risk (OARs) is crucial for maximizing target coverage while minimizing toxicities during radiotherapy [2,3]. Organs are normally contoured manually by experienced oncologists, a common clinical practice that entices three serious drawbacks: 1) it is labor-intensive and time-consuming to contour multiple organs slice-by-slice, 2) the diverse expertise and experience level of physicians introduce considerable intra-observer variations [3], and 3) low-contrast and fuzzy or ambiguous boundaries of organs on medical images can cause significant interobserver variation of delineation [4]. Time-consuming manual segmentation processes cannot support adaptive treatments when patients are on the table. Intra- and inter-observer variations introduce uncertainty to treatment planning that potentially compromises treatment outcomes. Automated segmentation approaches that rapidly contour organs with reliable and robust quality can overcome these drawbacks and provide accurate and efficient radiotherapy treatment.

Deep learning [5] has been evolving rapidly as an enabling technique for various real-world problems in the past decades. It has been extended and applied to medical image segmentation to provide accurate, reliable, and efficient delineation of pelvic CT images [2,3,6-13]. However, deep learning in its basic form of supervised learning requires large quantities of annotated data to train effective models, which presents a severe bottleneck for medical imaging applications. High-performance deep learning segmentation models that work on natural images are usually derived from massive data, e.g., the ImageNet dataset contains 14 million images [14]. However, unlike natural images, medical images need to be annotated by professionals with specific expertise, making data acquisition and annotation laborious, expensive, and time-consuming [6]. Moreover, inter-observer variation, caused by the difficulty of annotating and diverse expertise levels, imposes inconsistency on the dataset and compromises the learning outcome [6]. As a result, a common bottleneck in the medical imaging area is the lack of high-quality annotated datasets with a unified standard.

Here, we propose a novel semi-supervised adversarial deep learning approach to address this bottleneck by utilizing and synthesizing un-annotated data. Instead of using all the CT images and annotated contours for the training data, we use the unannotated CT images as part of our training data for the semi-supervised learning scheme. Although high-quality annotations are costly and often inadequate, raw data is often sufficiently collected in clinical practice. Furthermore, the new method can also synthesize new images to enlarge the set of un-annotated data, if needed, to enhance training and ultimately improve segmentation models. In short, the new approach has two prominent features. First, it adopts an adversarial learning scheme to utilize un-annotated data for training so that fewer annotated data are required. Second, the new approach can synthesize new data from a limited number of annotated images and support a semi-supervised adversarial learning scheme.

The generative adversarial network (GAN) [15], a special deep learning framework, has been extensively studied and applied to many real-world problems. A favorable feature of GAN is its ability to generate data with a desirable distribution [16]. GAN consists of two major components, a generator and a discriminator. The generator is trained to understand and generate synthetic data obeying certain distributions. The discriminator is tasked to learn to distinguish the generated data from real data [15]. The two components are trained in an adversarial scheme by which they compete against each other until reaching equilibrium – the generator attempts to produce synthetic data similar to the given examples to fool the discriminator.

In contrast, the discriminator tries to distinguish synthetic data from genuine ones. Several variations of GAN, such as conditional GAN (CGAN) [17], progressive growth GAN (PGGAN) [18], and triple GAN[19], have been proposed and tailored to specific applications. These generative methods perform well for data generation, which is particularly desirable when training examples are insufficient. GAN has been used to synthesize raw medical images, such as CTs or MRIs. However, such synthesized images are not new at all but rather transformed from images in a different format, e.g., from CT images to MRIs, as done in [20]. In other words, these synthesized MRIs are not new data points because they are just from the same views of the same cases but represented in different data formats. GAN-aided new case synthesis has also been explored for classification [21], but high-quality new case synthesis has not been explored much for segmentation. Generating new cases for segmentation requires a high representation ability of the generator, so we implemented a more advanced GAN structure and a more sophisticated training strategy.

Besides generating new data, the idea of adversarial learning lends itself to supervised learning. Luc *et al.* cast a network for segmentation as a generator and adopt the adversarial learning scheme for semantic segmentation, which performed well on the Pascal VOC 2012 dataset [22]. Hung *et al.* proposed a semi-supervised adversarial segmentation method using images with incomplete annotations for training [23].

To utilize GAN-augmented data without manual annotations, we developed a new semi-supervised deep learning approach for medical image segmentation. The new method has three major components: a CNN-based segmentation network (S-net), a discriminator network (D-net) for adversarial learning, and a progressive growth generative adversarial network (PGGAN). We adopted the residual U-net [24] as the backbone architecture for S-net and introduced several modifications to the overall design. We used an adversarial learning scheme, formulated in a generative adversarial network (GAN) [15], to train D-net to distinguish predicted label maps and ground-truth label maps. We trained a PGGAN for data augmentation [25] to synthesize new un-annotated data, when needed, from a small number of annotated training images for semi-supervised learning.

We applied the novel approach to 3D Pelvic CT segmentation as a case study to demonstrate its power on applications where training data are far from adequate. In our study, we also compared the new method and model with several state-of-the-art segmentation methods using three widely used metrics: dice similarity coefficient (DSC), average Hausdorff distance (AHD), and surface Hausdorff distance (ASHD).

## 2 Methods

### 2.1 Data Used and Augmentation

Institutional review board (IRB) approval was obtained for this study from Washington University of St. Louis. Planning CT and structure data sets for 120 intact prostate cancer patients were retrospectively selected. All CT images were acquired by a 16-slice CT scanner with an 85 cm bore size (Philips Brilliance Big Bore, Cleveland, OH, US).

The manually delineated contours served as ground truth in this study. The OARs were drawn by the same technician, and the prostate contours were drawn by two radiation oncologists with over ten years of experience treating prostate cancer patients and with a consensus contour generated using the Eclipse treatment planning system (Varian Medical Systems, CA). We fused the CT with MRI, which allowed for accurate delineation of the prostate. Individual lesions were not contoured as the standard clinical practice is to treat the entire prostate. Despite lesions that can sometimes be visualized in the MRI, prostate cancer tends to have a multifocal nature, and treating the entire gland is standard. Patients who had a prostatectomy were excluded from the study (Zhang et al.,2020).

From each of the 120 patients, we collected 100-200 slices with a slice thickness of 1.5mm. Every CT slice was translated into a 2D array with $512 \times 512$ pixels according to the original pixel spacing (0.97mm). We stacked 2D CT slices to build 3D CT volumes to cover the slice thickness. To accommodate the workstation used in the experiments, we first center-cropped each slice to the size of $384 \times 384$ and then resized them to $128 \times 128$ and picked the middle 64 slices from each case. Thus, we built a 3D volume with a dimension of $64 \times 128 \times 128$ for every patient.

We randomly split the 120 cases into a train/validation set of 100 cases and a test set of 20 cases. We used the train/validation set with 100 patients to perform 10-fold cross-validation, where 90 cases were used for training and 10 cases for validation each time. The test set (of 20 cases) was used for model evaluation and performance comparison. The size of the training set (of 90 cases) was small and may cause a serious overfitting problem, so we used a random sampling technique for data augmentation. For each training iteration, we randomly picked 16 continuous slices from the $64 \times 128 \times 128$ CT volumes and cropped the ground-truth label maps accordingly. Thus, each $64 \times 128 \times 128$ volume can generate 48 different $16 \times 128 \times 128$ volumes depending on different cropping positions. These smaller volumes may contain different OARs, and the positions of OARs were shifted between volumes cropped from nearby positions. This technique effectively augmented the training data by adding diversity and random translations to fixed organ positions. During the inference and testing phase, every validation/testing case was cropped into four 16-slice volumes and then fed into the model one by one. The outcomes from the network were later stacked together to restore the original dimensions ($64 \times 128 \times 128$) for evaluation.

### 2.2 Overview of the proposed semi-supervised learning method

To reiterate, our new method (Fig. 1) included three major components: segmentation network (*S-net*), discriminator network (*D-net*), and progressive growth GAN (*PGGAN* [18]). The method worked in two modes: 1) PGGAN-aided data augmentation; 2) semi-supervised adversarial learning scheme using *S-net* and *D-net*. These two modes of action were integrated to address the bottleneck problem of insufficient data. We first trained a CT synthesis PGGAN and synthesized un-annotated CT images, which we later used for semi-supervised learning. We then designed a semi-supervised adversarial learning scheme involving *S-net* and *D-net*. This scheme worked with both the annotated training data and additional data synthesized by PGGAN. In other words, the semi-supervised adversarial learning scheme combined annotated and un-annotated data to improve the quality of the final model.

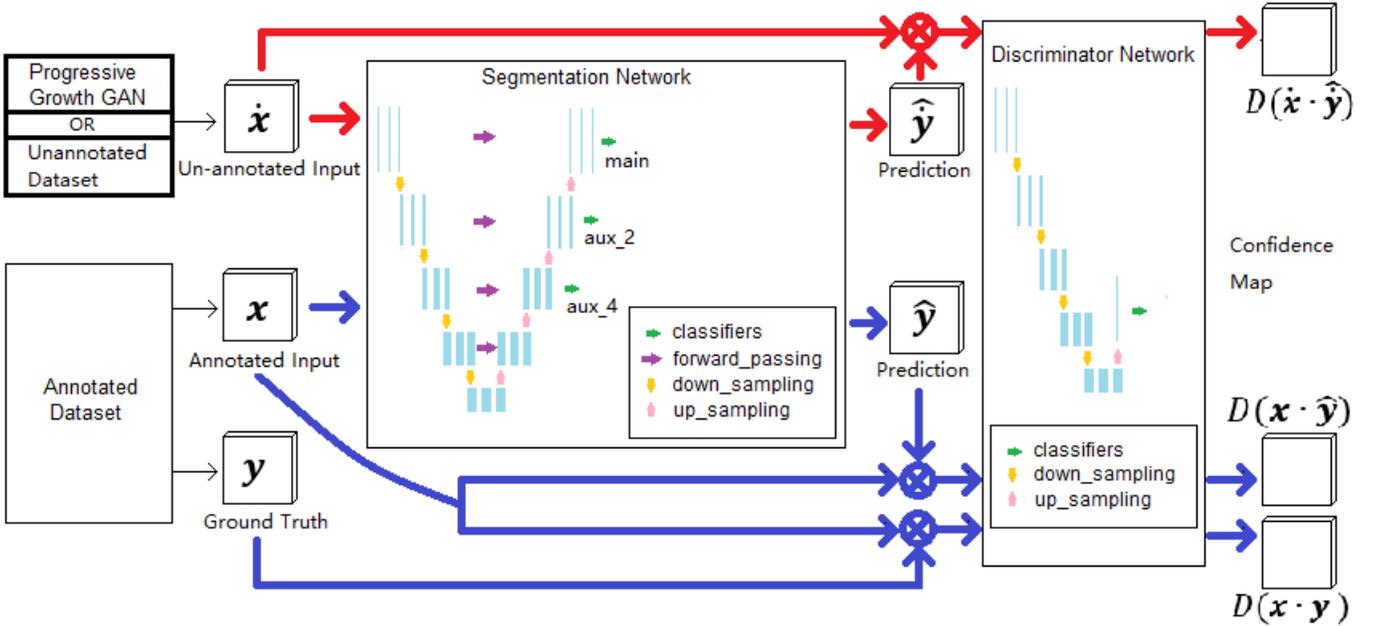

Fig. 1. Workflow of the proposed semi-supervised learning method. $D(\cdot)$ is the forward passing operation of the discriminator network.

The overall approach was designed for three scenarios: (1) when data annotations are unavailable, yet un-annotated data are available, it can utilize un-annotated data for training with the semi-supervised adversarial learning scheme; (2) when the given annotated training examples were insufficient, and un-annotated data were also unavailable, it can synthesize un-annotated data using PGGAN; and (3) when training examples were sufficient for learning, it can generate and use additional synthesized un-annotated data to improve the quality of training and final models by semi-supervised learning.

**2.3 Segmentation and discriminator network structure**

The segmentation network *S-net* adopted the residual U-net [1] as its backbone architecture. The residual connection allowed the feature maps to bypass the non-linear transformations with an identity mapping. This design went beyond regular skip connections and reformulated the layers as learning residual functions [26]. The residual U-net in *S-net* took $16 \times 128 \times 128$ CT volumes (16 slices of 128 by 128 CT scans) with one channel as its input. It output $16 \times 128 \times 128$ segmentation volumes with six channels, one for each of the OARs and the background. With the original U-net [24] architecture, *S-net* first down-sampled the input for feature extraction, then up-sampled the extracted feature maps to scale back to the original size and perform classification. During the up-sampling process, feature maps from the down-sampling path were passed for concatenation as guidance for high-resolution spatial context information [27]. Along the down-sampling path of the original U-net, multiple stride-2 max-pooling layers were used to shrink feature maps, which caused spatial information loss [28]. To preserve spatial information, we replaced all the stride-2 max-pooling layers with multi-scale pooling layers developed in our adversarial multi-residual multi-scale pooling Markov Random Field enhanced network (ARPM-net) [29].

Besides the classifier on the top level (in green in Fig. 1) that classified feature maps in the original resolution ($16 \times 128 \times 128$), *S-net* also included two auxiliary classifiers. The auxiliary classifier *aux_2* (Fig. 1) worked on $8 \times 64 \times 64$ feature maps with 128 channels, and the *aux_4* (Fig. 1) processed $4 \times 32 \times 32$ feature maps with 256 channels. The segmentation outputs of these two auxiliary classifiers were up-sampled by interpolation to rescale to the input dimension of $16 \times 128 \times 128$. This architecture design incorporated the compressed feature maps in the final prediction, allowing supervision learning in the latent space [30]. In the original residual U-net [1], there was only one classifier in the last layer of the network, and its outputs determined the loss from the ground truth. When the loss was back-propagated into the network, the supervision was only provided in the last layer. However, with two auxiliary classifiers contributing to the final prediction, supervision was done not only in the last layer but also in the middle layers. Thus, we introduced supervision into the latent space by the two auxiliary classifiers. The segmentation predictions of *aux_4* and *aux_2* were weighted by 0.25 and 0.5 in the final label prediction. The final label map predictions were in $16 \times 128 \times 128$ volumes, with six channels representing each voxel's probability of belonging to each of the six classes.

The discriminator network *D-net* implemented a similar structure as the encoder part of *S-net*. It contained four down-sampling layers, one classification layer, and one up-sampling layer (Fig. 1). In *D-net*, we also used residual convolution backbones as used in *S-net*. The input for *D-net* is not the 3D segmentation prediction mask from *S-net*. Instead, it is the voxel-wise product of input volumes and prediction masks. *D-net*'s output is a confidence map instead of numerous values representing real (1) or fake (0). *D-net* predicts the probability of input being computed from ground-truth masks or predicted masks for each voxel.

## 2.4 Semi-supervised Adversarial Training

Adversarial training is mainly used for GAN-based data generation [15]. It can potentially improve the original task's performance by using additional, albeit synthesized, data generated from random noises and training examples [20,22,29,31]. Semantic segmentation can be cast as a generative task. Rather than generating new images, *S-net* in our approach produced segmentation masks that closely resemble manually delineated masks on input CT images.

We added an adversarial learning scheme to the new approach by exploiting two significant merits of adversarial models. Firstly, without *D-net*, *S-net* was optimized based only on a cross-entropy loss, an element-wise (i.e., pixel- or voxel-wise) loss metric widely used for semantic segmentation [14,27,32]. Furthermore, we introduced a style-loss calculated by *D-net* into the training scheme to help *S-net* generate contours closely resemble manual contours. Secondly, like other CNN networks, *S-net* was designed for supervised learning so that it cannot make use of un-annotated CT images. Thanks to the adversarial learning scheme, *S-net* can be extended to learning from un-annotated CT images with the discriminator's help.

The basic adversarial training scheme had four steps:

1. Feed the input CT image $x$ forward through *S-net* to obtain predicted mask $\hat{y} = S(x)$, where $S(\cdot)$ represents the forward passing operation of *S-net*.
2. Feed the voxel-wise product of $x \cdot \hat{y}$ (fake input) or $x \cdot y$ (real input) into *D-net* to obtain a confidence map $D(x \cdot \hat{y})$ or $D(x \cdot y)$, respectively, where $y$ is the ground truth label map of CT image $x$, and $D(\cdot)$ represents forward passing operation of *D-net*.
3. Compute the losses for *S-net* and *D-net*. The main objective of adversarial learning is to train *S-net* to generate "fake" segmentation maps that closely resemble ground-truth label maps and can fool *D-net* and simultaneously train *D-net* to distinguish $x \cdot \hat{y}$ (fake input) from $x \cdot y$ (real input).
4. Backpropagate the losses for *S-net* and *D-net*. The loss for *S-net* (denoted as $\mathcal{L}_S$) has two components, i.e., 1) voxel-wise loss (denoted as $\mathcal{L}_{vox}$) between prediction $\hat{y}$ and ground-truth $y$, and 2) adversarial loss (denoted as $\mathcal{L}_{adv}$) computed interactively with *D-net*. The loss for *D-net* (denoted as $\mathcal{L}_D$) is computed based on how well the discriminator can separate fake inputs from real ones.

Besides the essential adversarial learning, we also utilized un-annotated CT images as input to S-net. Un-annotated CT images came from either real CT images without annotation or PGGAN-synthesized CT images. Similarly, we fed the un-annotated CT image $\dot{x}$ to S-net to predict $\hat{y} = S(\dot{x})$. Since there was no ground-truth $\dot{y}$ for $\dot{x}$, we cannot compute $\mathcal{L}_{vox}$ with un-annotated CT images, but we can use D-net to compute $D(\dot{x} \cdot S(\dot{x}))$ and take advantage of semi-supervised adversarial learning.

### 2.5.1 Training the Segmentation Network (S-net)

Inspired by Hung's work [23], *S-net* was trained with three losses: voxel-wise loss $\mathcal{L}_{vox}$, adversarial loss $\mathcal{L}_{adv}$, and semi-supervised learning loss $\mathcal{L}_{semi}$. Start with the voxel-wise loss $\mathcal{L}_{vox}$, adopting the adaptively weighted loss function from ARPM-net [29], the voxel-wise loss function (for each sample) of *S-net* is:

$$\ell^*_{mce}(\hat{y}, y) = -\sum_{Z \times H \times W} \sum_{c=1}^{C} w_c y_c \ln \hat{y}_c, \quad (1)$$

$$w_c = 2 - DSC_c + \ln \frac{number\ of\ voxel\ of\ all\ classes}{number\ of\ voxel\ of\ class\ c}. \quad (2)$$

In Eq. 1, $\ell^*_{mce}(\hat{y}, y)$ is the adaptively weighted multi-class cross-entropy (MCE) loss between the prediction $\hat{y}$ and ground-truth $y$; $Z$, $H$, and $W$ are the depth, height, and width of the 3D CT volume, respective; $C$ represents the number of classes and in our case $C = 6$; and $w_c$ is the adaptive weight of class $c$. The adaptive weight is calculated by Eq. 2, where $DSC_i$ is the current performance (measured by the Dice Similarity Coefficient) for class $i$. Combined, we have the voxel-wise loss $\mathcal{L}_{vox}$ for all $N$ samples as:

$$\mathcal{L}_{vox} = \sum_{n=1}^{N} \ell^*_{MCE}(S(x_n), y_n). \quad (3)$$

The adversarial learning trained *S-net* to generate segmentation predictions that can fool *D-net* to classify a synthetic CT image as a real image. The adversarial loss $\mathcal{L}_{adv}$ measures the difference between the current *S-net* and a "perfect generator" that always fools *D-net*. The loss can be written as:

$$\ell_{adv} = \ell_{bce}(D(x_n \cdot S(x_n)), \mathbf{1}), \quad (4)$$

where **1** represents the target confidence map of *D-net* with value 1 (real) for all voxels; $\ell_{bce}(\cdot)$ is the binary cross-entropy (BCE) loss function:

$$\ell_{bce}(\hat{z}, z) = -\sum_{i=1}^{Z \times H \times W} [z_i \ln \hat{z}_i + (1 - z_i) \ln(1 - \hat{z}_i)], \quad (5)$$

$$\hat{z} = D(x_n \cdot S(x_n)), \quad (6)$$

$$z = 1. \quad (7)$$

Thus, we have the adversarial loss $\mathcal{L}_{adv}$ for all $N$ samples:

$$\mathcal{L}_{adv} = -\sum_{n=1}^{N} \sum_{Z \times H \times W} \ln(D(x_n \cdot S(x_n))). \quad (8)$$

In semi-supervised learning, un-annotated (or synthetic) CT images can be utilized for *S-net* training. Since there is no ground-truth available, we cannot compute $\mathcal{L}_{vox}$ for un-annotated data. However, for an un-annotated image $\dot{x}$, the trained discriminator *D-net* generates a confidence map $D(\dot{x} \cdot S(\dot{x}))$, which can infer the regions that are sufficiently close to ground-truth label maps [23]. These regions are selected based on a pre-set threshold $T_{semi}$. For image $\dot{x}$, we also have the self-taught "ground-truth" $\tilde{y}$ as an element-wise set with $\tilde{y}^{(z,h,w,c^*)} = 1$ if $c^* = argmax_c S(x)^{(z,h,w,c)}$, where $z, h, w$, and $c$ denote the voxels at the location $(z, h, w)$ and channel $c$, respectively. Thus, the semi-supervised loss can be written as:

$$\mathcal{L}_{semi} = -\sum_{n=1}^{N} \sum_{Z,H,W} \sum_{c=1}^{C} I^{(z,h,w)} \cdot \tilde{y}_n^{(z,h,w,c)} \ln(S(\dot{x}_n)^{(z,h,w)}), \quad (9)$$

$$I^{(z,h,w)} = \left( D(\dot{x}_n \cdot S(\dot{x}_n))^{(z,h,w)} > T_{semi} \right). \quad (10)$$

The indicator function $I^{(z,h,w)}$ in Eq. 10 indicates if *S-net* prediction on voxel at $(z,h,w)$ is sufficiently trustworthy, and threshold $T_{semi}$ controls the sensitivity [23]. We also pre-set the threshold to be 0.2, as suggested in [23]. The total loss function $\mathcal{L}_{S-net}$ is a weighted summation of all losses:

$$\mathcal{L}_S = \mathcal{L}_{vox} + \lambda_{adv}\mathcal{L}_{adv} + \lambda_{semi}\mathcal{L}_{semi}. \tag{11}$$

**2.5.2 Training the Discriminator Network (D-net)**

We train *D-net* as a competitor against *S-net* in adversarial learning [15]. The discriminator learns to separate the label map prediction generated by *S-net* from ground-truth. For input CT image $x_n$ with its ground-truth label map $y_n$, we use the binary cross-entropy loss function:

$$\mathcal{L}_D = -\sum_{n=1}^{N}[\ell_{bce}(D(x_n \cdot y_n),1) + \ell_{bce}(D(x_n \cdot S(x_n)),0)]. \tag{12}$$

*D-net* learns to label all voxels on the confidence map as 1 (real) if the input is the product computed with the ground-truth and 0 (fake) with *S-net* prediction. In Hung's work, they did not encounter the issue that we encountered, i.e., *D-net* easily distinguishes whether the probability maps came from the ground truth by detecting the one-hot probability [22,23]. We resolved this issue using the element-wise product of the input CT image and the one-hot encoded label prediction used as the input for *D-net*.

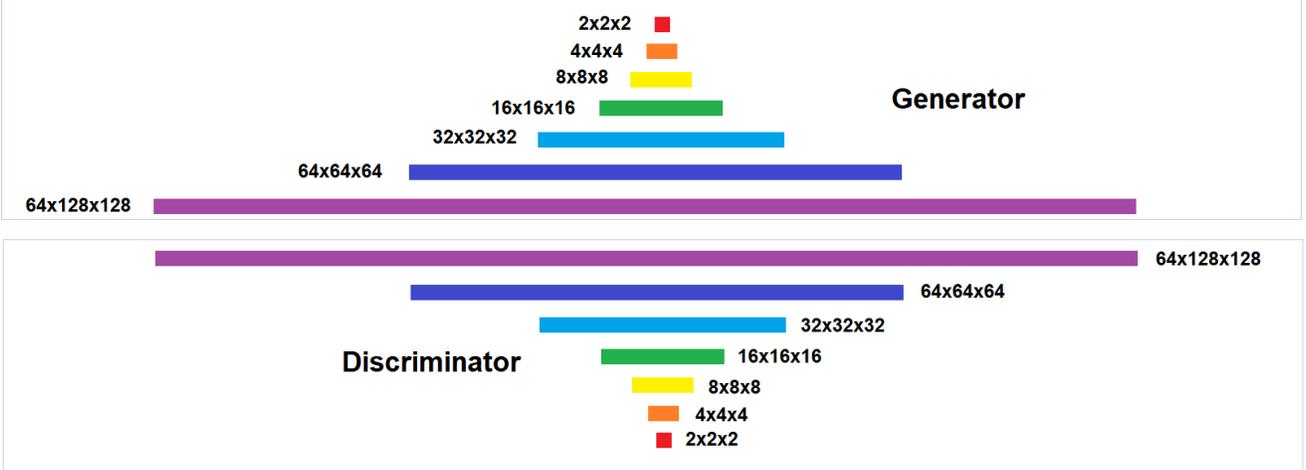

Fig. 2 Structure of the progressive growth generative adversarial network (PGGAN)

**2.6 Progressive Growth GAN-aided CT Synthesis**

The generative adversarial network can generate synthetic samples from high-dimensional data distributions [16]. The general framework supports the generation of different data types, and different GAN variations work specifically well on certain data types. The progressive growth generative adversarial network (PGGAN), as a special GAN, has been tailored to image synthesis [18]. PGGAN incrementally generates images with increasingly higher resolutions. It first discovers large image distribution structures and then shifts attention to increasingly finer details [18].

We adopted the training scheme of PGGAN and designed a new network architecture for synthesizing 3D CT volumes (Fig. 2). The generator $\mathcal{G}$ and the discriminator $\mathcal{D}$ had mirrored architectures. The training started by generating $4 \times 4 \times 4$ volumes. $\mathcal{G}$ took a $2 \times 2 \times 2$ random noise $z$ as its input and output a $4 \times 4 \times 4$ synthesized volume $x'$. $\mathcal{D}$ took $x'$ or down-sampled real CT volume $x$ as its input and output a prediction that the input was real ($x$) or fake ($x'$). After $\mathcal{G}$ was well-trained on low-resolution ($n \times n \times n$) volumes, one more up-sampling block was added to $\mathcal{G}$ and $\mathcal{D}$ individually to generate $2n \times 2n \times 2n$ volumes. This process repeated until the last up-sampling blocks were added to generate $64 \times 128 \times 128$ from $64 \times 64 \times 64$ feature maps. Note that $64 \times 128 \times 128$ synthesized CT volumes had the same dimensions as real CT volumes, and both were cropped to $16 \times 128 \times 128$ before being fed into *S-net*. Notice that adding randomly initiated layers to well-trained $\mathcal{G}$ and $\mathcal{D}$ could be disruptive to reach a training equilibrium. To address this issue, we used the smooth fade-in trick proposed in [18] to linearly increase weight $\alpha$ to combine the outputs from an already trained low-resolution layer and a newly added higher-resolution layer. We used the well-trained PGGAN to synthesize 30 CT volumes to aid semi-supervised learning in the subsequent adversarial semantic segmentation training.

## 2.7 Implementation and training

We implemented the new method in PyTorch and trained the overall model on two RTX 2080 Ti GPUs using data parallelism. For *S-net*, we used Adam optimizer [33] with an initial learning rate of $5 \times 10^{-4}$ and polynomial learning rate scheduler with a power of 0.9. For *D-net*, we used Adam optimizer with an initial learning rate of $10^{-4}$ and the same learning rate scheduler as for *S-net*. For the other parameters in Eq. 11, we set $\lambda_{adv} = 0.01 \; or \; 0.001$ for annotated and un-annotated data, respectively, and $\lambda_{semi} = 0.1$ as suggested in [23].

In the semi-supervised adversarial learning scheme, we used both annotated and un-annotated data (*config.B* and *config.C*) for training. Following [23], we pre-trained *S-net* and *D-net* with only labeled data for 5000 iterations before semi-supervised training started. The pre-training iterations can help stabilize the randomly initiated model. After pre-trained *S-net* and *D-net*, we randomly interleaved annotated and un-annotated data for semi-supervised training [23]. We trained the semi-supervised adversarial model on *config.B* and *config.C* dataset for 40k iterations with batch size 2. We also implemented and trained several state-of-the-art existing methods to derive some baseline models for model comparison and evaluation (Table II): 1) *3d_res_Unet* is a vanilla residual U-net model as proposed in [1]; 2) *3d_res_Unet_aux* is the *res_Unet* upgraded with auxiliary classifiers; 3) *3d_res_Unet_aux_adv* is the adversarial learning version of *3d_res_Unet_aux*; and 4) *3d_res_Unet_aux_adv_semi* is the complete version of our proposed method. The first three models can be trained with *config.A*, and only the complete version can be built with *config.B* and *config.C*. Experiments 1, 2, and 3 used the same training data, 90 annotated cases. These three experiments were designed to test the improvement provided by the auxiliary classifiers and the supervised adversarial learning scheme. Experiment 5 was performed on 60 annotated cases randomly selected from the 90 annotated cases available. We used the rest 30 cases as unannotated data for semi-supervised adversarial learning. The experiment was designed to test the model's ability to learn from real unannotated CT images when we have enough unannotated data available. Experiment 6 was performed on 90 annotated cases with extra 30 synthesized cases, which was designed to test the model's ability to learn from synthesized unannotated CT images when we want to augment the dataset for better learning outcomes.

Considering the small size of the dataset, we applied traditional data augmentation to all input volumes during the training phase [24] to alleviate the overfitting issue. Data augmentation techniques included random translation (shifting the volume along X or Y-axis), random sampling (selecting slices at a random position along the Z-axis), randomly flipping (mirroring the volumes left to right), adjusting contrast, and random rotation (rotating the volume by -10 to 10 degrees). These traditional data augmentation techniques were applied to all experiments, while GAN-aided data synthesis was only used for semi-supervised learning.

## 2.8 Evaluation Metrics

Four evaluation metrics were used in our experiments: dice similarity coefficient [34], average Hausdorff distance [35], average surface Hausdorff distance [36], and relative volume difference [36]. Dice similarity coefficient (DSC) is a widely used segmentation metric in medical imaging:

$$DSC = \frac{2 \times |True \cap Pred|}{|True| + |Pred|}, \tag{13}$$

where $|True|$ and $|Pred|$ are the number of voxels in the ground-truth and prediction, respectively [29]. The average Hausdorff distance (AHD) [3] measures the maximal average point-wise distance from points in $X$ to the nearest point in $Y$, and the average point-wise distance from points in $Y$ to the nearest point in $X$:

$$AHD(X,Y) = max(\frac{1}{|X|}\sum_{x \in X} \min_{y \in Y} d(x,y), \frac{1}{|Y|}\sum_{y \in Y} \min_{x \in X} d(y,x)), \tag{14}$$

where $X$ and $Y$ are the voxel sets of the ground-truth and prediction volume, respectively, and $d(x,y)$ is the Euclidean distance from point $x$ to point $y$. The Average Surface Hausdorff Distance (ASHD) [3] is the symmetrical average point-wise distance between a point on one surface to the nearest point on the other surface:

$$ASHD(X,Y) = \frac{1}{2}(\frac{1}{|X|}\sum_{x \in X} \min_{y \in Y} d(x,y) + \frac{1}{|Y|}\sum_{y \in Y} \min_{x \in X} d(y,x)), \tag{15}$$

where $X$ and $Y$ are the voxel sets of the ground-truth and prediction surface, respectively, and $d(x,y)$ is the Euclidean distance from point $x$ to $y$.

## 3. Results

### 3.1 Evaluation of Progressive Growth GAN-aided CT Synthesis

Fig. 3 shows the examples of two synthesized cases, each showing five slices at different levels exhibiting all OARs. We hand-picked 30 synthesized cases with the same medical significance as real clinical CT images and calculated the average pixel/voxel intensity for each organ. We compared results to those calculated based on the 90 ground-truth cases. The Hounsfield Units (HU) difference for each organ is shown in Table I. The quality of these images has been reviewed and confirmed by one of the coauthors, a clinician and radiation oncologist with over ten years of experience. The synthesized images show a reasonably good quality on bladder, rectum, and Femurs. It should be noted that the prostate cannot be easily delineated even in real CT scans. The quality of our synthesized images is comparable to earlier research in terms of HU differences between the ground truth and synthesized images.[37]

| Organ | Prostate | Bladder | Rectum | Femur_L | Femur_R |
|---|---|---|---|---|---|
| HU error (mean STD) | 7.5 ± 19.7 | 5.4 ± 35.1 | 17.9 ± 44.78 | 3.8 ± 122.83 | 4.1 ± 110.37 |

Table I: HU error (mean ± standard deviation) for each OAR. Note that positive difference means synthesized CTs have a higher intensity than real CTs.

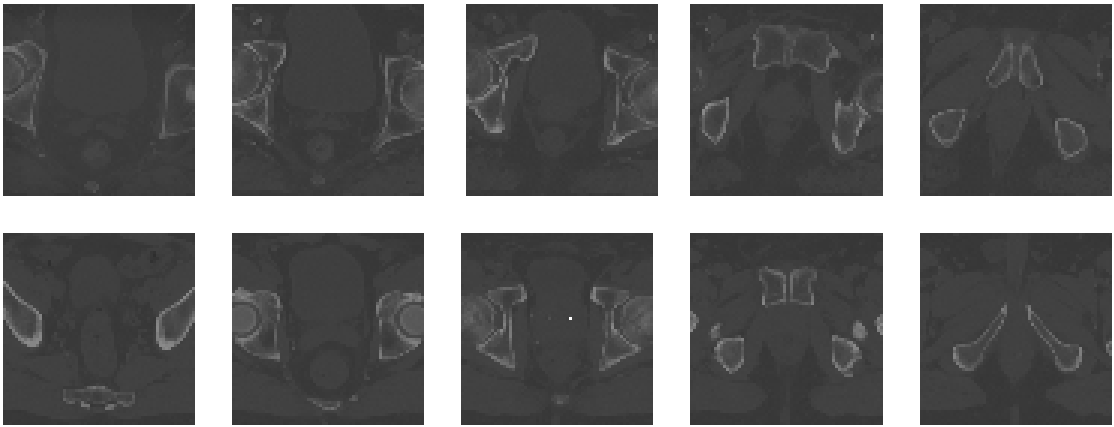

Fig. 3  Examples of two PGGAN-synthesized cases. We show five slices for each case to exhibit the quality of organs.

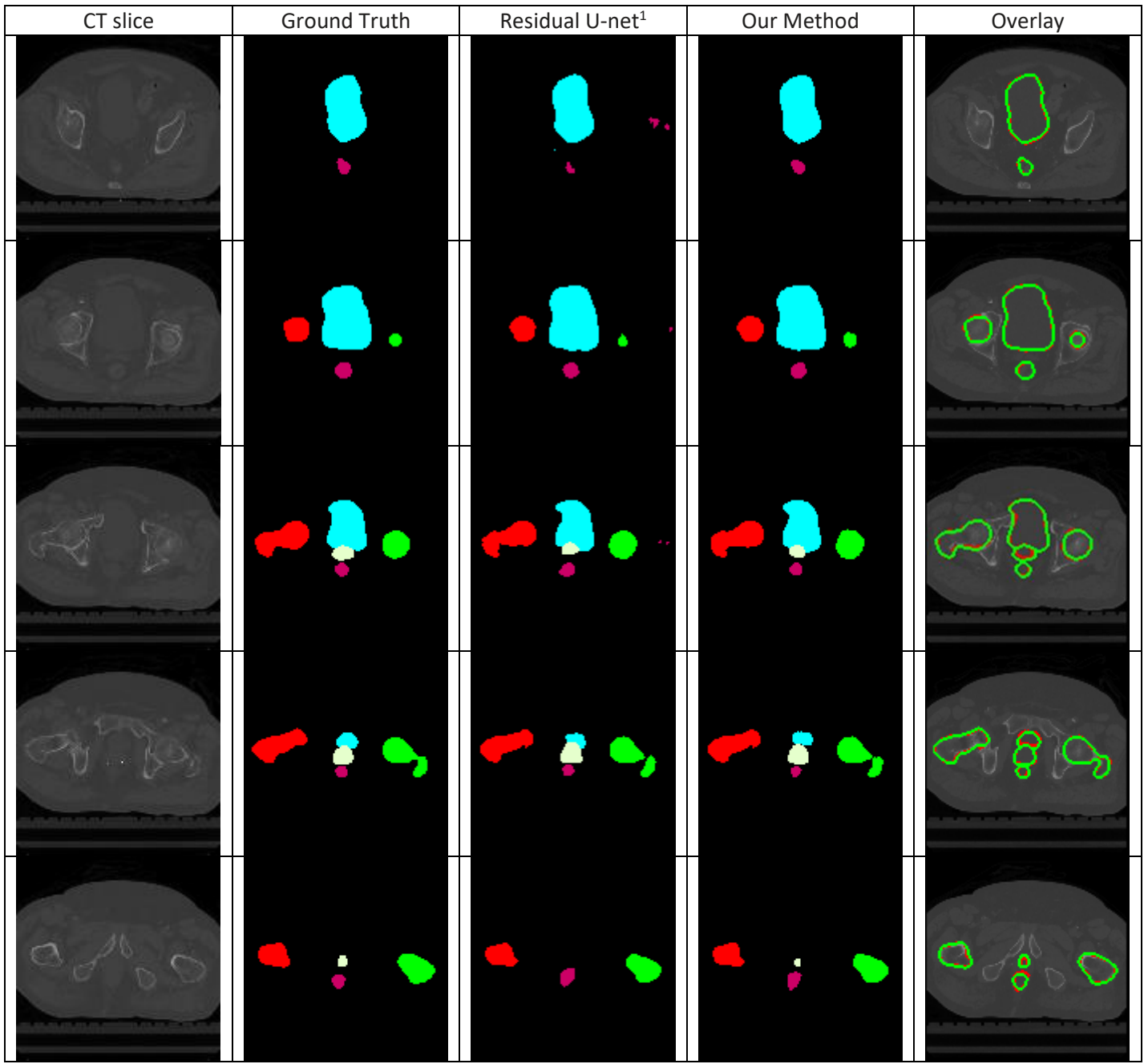

Fig. 4. Exhibition of segmentation results. Five columns are CT slices, ground-truth (manual annotations by experts) label masks (bladder in light blue, prostate in beige, rectum in purple, and femurs in red and green), prediction label masks of baseline method, prediction label masks of our method, and contour overlay (ground truth in green and prediction in red) of the proposed method. The five slices are selected from the same test case and ordered from superior to inferior.

### 3.2 Model Evaluation

We evaluated the performance of our new method and compared it with the baseline methods by 10-fold cross-validation. The new approach achieved nearly 100% accuracy on both femurs (Table II) thanks to the high contrast of the CT images analyzed. However, the prostate, bladder, and rectum were more challenging to segment due to their low contrast and fuzzy boundaries. Despite the low contrast, our approach (Experiment 6) generates high-quality segmentation contours closely aligned with the ground truth (Figure

4). Compared to manual segmentation that typically takes 20-30 minutes per patient for the prostate alone, our approach can segment five organs in 10 seconds per patient. The results suggest that the new approach can improve the quality of routine clinical practice by rapidly producing accurate contours with high efficiency.

### 3.3 Effectiveness of Semi-supervised Adversarial Learning

An eminent feature of the new method is its ability to use both annotated and un-annotated images. The lack-of-data bottleneck can be adequately dealt with when the model can learn from unannotated data, guaranteed by sufficient clinical data and our backup plan of data synthesis. The new model is meritorious in multiple ways. If annotated data are expensive to collect but unannotated data are sufficient, the model can use un-annotated data by taking advantage of the data generation capability of adversarial learning. Moreover, even when raw data are not sufficient, the model can produce synthesized data using adversarial learning of GAN. When enough annotated data are available, we can always generate more raw data for data augmentation.

We compared the proposed approach and the baseline methods to show that the former achieves comparable results with less annotated data and better results with the same amount of annotated data (Table II).

| Experiment No. | Model | Dataset | Prostate DSC AHD(mm) ASHD(mm) VD(%) | Bladder DSC AHD(mm) ASHD(mm) VD(%) | Rectum DSC AHD(mm) ASHD(mm) VD(%) | Femur_L DSC AHD(mm) ASHD(mm) VD(%) | Femur_R DSC AHD(mm) ASHD(mm) VD(%) |
|---|---|---|---|---|---|---|---|
| 1 | *3d_res_Unet* [30] | 90 labeled | 0.85(±0.12) 0.34(±0.13) 0.80(±0.17) -11.13 | 0.95(±0.10) 0.12(±0.10) 0.50(±0.23) -0.98 | 0.84(±0.12) 0.40(±0.46) 0.68(±0.31) -7.72 | 0.97(±0.01) 0.06(±0.04) 0.27(±0.16) -1.62 | 0.97(±0.01) 0.04(±0.01) 0.18(±0.05) -1.89 |
| 2 | *3d_res_Unet_aux* | 90 labeled | 0.86(±0.14) 0.33(±0.16) 0.78(±0.17) -4.27 | 0.95(±0.07) 0.10(±0.10) 0.42(±0.17) +2.92 | 0.86(±0.9) 0.39(±0.32) 0.65(±0.30) -1.52 | 0.97(±0.01) 0.04(±0.02) 0.25(±0.10) -1.24 | 0.97(±0.01) 0.04(±0.02) 0.25(±0.18) -1.27 |
| 3 | *3d_res_Unet_aux_adv* | 90 labeled | 0.87(±0.15) 0.31(±0.16) 0.77(±0.21) -3.07 | 0.96(±0.07) 0.09(±0.08) 0.41(±0.14) +3.22 | 0.87(±0.13) 0.36(±0.35) 0.63(±0.27) -0.97 | 0.97(±0.01) 0.04(±0.02) 0.22(±0.09) -0.53 | 0.97(±0.01) 0.06(±0.07) 0.23(±0.16) -0.97 |
| 4 | *3d_res_Unet_aux_adv* | 60 labeled | 0.84(±0.18) 0.45(±0.26) 0.86(±0.17) -5.10 | 0.92(±0.20) 0.73(±0.50) 0.80(±0.44) -2.24 | 0.83(±0.23) 0.53(±0.54) 0.72(±0.42) -1.35 | 0.95(±0.02) 0.06(±0.01) 0.23(±0.10) +1.22 | 0.96(±0.02) 0.07(±0.01) 0.24(±0.04) -1.34 |
| 5 | *3d_res_Unet_aux_adv_semi* (our method) | 60 labeled & 30 unlabeled | 0.86(±0.13) 0.36(±0.20) 0.82(±0.15) -4.45 | 0.96(±0.07) 0.12(±0.11) 0.51(±0.17) +3.24 | 0.86(±0.12) 0.36(±0.34) 0.66(±0.31) -1.01 | 0.97(±0.01) 0.05(±0.03) 0.21(±0.08) -0.66 | 0.97(±0.01) 0.06(±0.08) 0.22(±0.11) -0.99 |
| 6 | *3d_res_Unet_aux_adv_semi* (our method) | 90 labeled & 30 unlabeled | 0.90(±0.09) 0.23(±0.11) 0.62(±0.15) -4.38 | 0.96(±0.06) 0.12(±0.11) 0.45(±0.22) +0.80 | 0.87(±0.11) 0.28(±0.17) 0.57(±0.18) -5.46 | 0.97(±0.01) 0.04(±0.02) 0.22(±0.09) -0.01 | 0.97(±0.01) 0.04(±0.02) 0.19(±0.07) -0.64 |

Table II: Experiment results from 10-fold cross-validation. We evaluated the performance with four metrics: dice similarity coefficient (DSC), average Hausdorff distance (AHD), average surface Hausdorff distance (ASHD), and relative volume difference (VD) on five OARs: prostate, bladder, rectum, left femur (Femur_L) and right femur (Femur_R). The best score of each metric on each organ is colored in red.

Experiments 1 and 2 (see Methods) were designed to measure the effectiveness of auxiliary classifiers. Even though the prostate and rectum are challenging to segment due to their low contrast level and variations in size and shape, using auxiliary classifiers in the new method can improve the quality of the prostate (reducing Dice by 7.0% from 0.86 to 0.8) and rectum contours (reducing Dice by 2.4% from 0.86 to 0.84). With auxiliary classifiers that perform segmentation on feature maps of multiple scales, the model generated better contours for organs of various shapes and sizes. Experiments 2 and 3 were used to analyze the effectiveness of the adversarial training scheme. The results showed that the performance on the prostate and rectum was further improved with adversarial training, which illustrated the effectiveness of adding style-wise loss into the loss function.

We designed experiments 3, 4, and 5 to test if our method can achieve comparable or even better performance with less annotated data. The results showed that discarding one-third of the annotated examples would not significantly degrade the performance with the new semi-supervised learning algorithm. As shown in Table II, discarding one-third of the annotated data would only decrease the Dice by 1% on the prostate and rectum. With Experiments 3-5, we demonstrated that the new method required less effort for data annotation.

We also expected additional synthesized CT images to improve the performance with the same amount of annotated data. Thus, we introduced experiments 3 and 6 to test if our method can achieve better performance with PGGAN-synthesized data augmentation. Indeed, the combination of using semi-supervised loss and synthesized data produced by adversarial learning was effective, providing comparable or better results (Table II).

We tested if our semi-supervised learning scheme was robust with diverse types of input data. In experiment 5, the un-annotated cases used in training were real CT images without annotations. However, in experiment 6, the 30 un-annotated training cases were synthesized data. The results from experiments 5 and 6 showed that PGGAN-aided CT synthesis was effective. More importantly, both real un-annotated images (Exp. 5) and PGGAN-synthesized un-annotated images (Exp. 6) supported well the semi-supervised in our new method.

Table III: Model performance on testing data. The best score of each metric on each organ is colored in red. Note that relative volume difference is not listed because the metric was not used in most of compared methods.

| | Prostate DSC AHD(mm) ASHD(mm) | Bladder DSC AHD(mm) ASHD(mm) | Rectum DSC AHD(mm) ASHD(mm) | Femur_L DSC AHD(mm) ASHD(mm) | Femur_R DSC AHD(mm) ASHD(mm) |
|---|---|---|---|---|---|
| Multi-atlas Based Segmentation (Acosta et al., 2014) | 0.85(±0.004) - - | 0.92(±0.002) - - | 0.80(±0.007) - - | N/A - - | N/A - - |
| Deep Dilated CNN (Men et al., 2017) | 0.88 - - | 0.93 - - | 0.62 - - | 0.92 - - | 0.92 - - |
| 2D U-net (Kazemifar et al., 2018) | 0.88(±0.12) 0.4(±0.7) 1.2(±0.9) | 0.95(±0.04) 0.4(±0.6) 1.1(±0.8) | 0.92(±0.06) 0.2(±0.3) 0.8(±0.6) | N/A | N/A |
| 2D U-net Localization, 3D U-net Segmentation (Balagopal et al., 2018) | 0.90(±0.02) 5.3(±2.8) 0.7(±0.5) | 0.95(±0.02) 17.0(±14.6) 0.5(±0.7) | 0.84(±0.04) 4.9(±3.9) 0.8(±0.7) | 0.96(±0.03) - - | 0.95(±0.01) - - |
| 3d_res_Unet_aux_adv_semi (our method) | 0.90(±0.12) 0.27(±0.13) 0.77(±0.20) | 0.95(±0.05) 0.11(±0.13) 0.47(±0.23) | 0.87(±0.10) 0.30(±0.19) 0.63(±0.21) | 0.97(±0.01) 0.04(±0.01) 0.24(±0.11) | 0.97(±0.01) 0.04(±0.02) 0.18(±0.06) |

## 4. Discussion

We compared our model's performance on test data with several state-of-the-art methods for pelvic CT segmentation, including one multi-atlas based model [4] and four deep learning-based models. Deep Dilated CNN [11] and 2D U-net [3] are end-to-end models, and 2D U-net Localization plus 3D U-net Segmentation [2] is a two-step region-of-interests (ROI) segmentation approach. Since we had no access to their datasets nor source code, we tried our best to implement and train their models on our dataset yet did not achieve the same performance. Hence, we did not compare these methods directly on the same dataset. Instead, we accepted their reported model performances as their best results (Table III).

The proposed method outperformed the multi-atlas based model [4] and the Deep Dilated CNN method [11] with higher DSCs on all OARs. The 2D U-net and 2D U-net Localization plus 3D U-net Segmentation are ROI segmentation methods. Multiple networks are trained for different OARs catering to different shapes, sizes, and contrast levels. The method required one 2D localization network and four 3D segmentation networks with different architectural designs to segment four different OARs. Our method achieved similar DSCs on all OARs as the two ROI segmentation methods. ROI segmentation methods need to crop off volumes for each organ, segment them separately with multiple networks, then merge their label masks. However, our model performed an end-to-end segmentation process and segmented all organs with one trained network. We do not need to generate or store intermediate volumes or post-processing of the model output, costly and hard to implement. With the proposed new model, the average processing time for a patient with 100-200 CT slices is less than 10 seconds, which is a huge speed-up compared to "a few minutes" required by the two-step method [2].

Table IV: Model performance on testing data. The best score of each metric on each organ is colored in red. Statistically significant (p-value < 0.05) improvements are marked with '*'.

|  | Prostate DSC AHD(mm) ASHD(mm) | Bladder DSC AHD(mm) ASHD(mm) | Rectum DSC AHD(mm) ASHD(mm) | Femur_L DSC AHD(mm) ASHD(mm) | Femur_R DSC AHD(mm) ASHD(mm) |
|---|---|---|---|---|---|
| ARPM-net (Zhang et al., 2020) | 0.88(±0.11) 1.58(±1.77) 2.11(±2.03) | 0.97(±0.07) 1.91(±1.29) 2.36(±2.43) | 0.86(±0.12) 3.14(±2.39) 3.05(±2.11) | 0.97(±0.01) 1.76(±1.57) 1.99(±1.66) | 0.97(±0.01) 1.92(±1.01) 2.00(±2.07) |
| 3d_res_Unet_aux_adv_semi (our method) | 0.90(±0.12) 0.27(±0.13) 0.77(±0.20) | 0.95(±0.05) 0.11(±0.13) 0.47(±0.23) | 0.87(±0.10) 0.30(±0.19) 0.63(±0.21) | 0.97(±0.01) 0.04(±0.01) 0.24(±0.11) | 0.97(±0.01) 0.04(±0.02) 0.18(±0.06) |
| p-value | <0.01* | <0.01* | <0.01* | 0.111 | 0.125 |

Compared to our recent study using ARPM-net, which also segmented all OARs within one forward propagation [29], our new method achieved better DSCs, AHDs, and ASHDs on the prostate and rectum (Table IV). Our new method was trained and tested with the same dataset as ARPM-net. Our semi-supervised adversarial learning scheme supported by PGGAN-aided CT synthesis had significantly smaller HDs and AHDs. Both methods leveraged the power of adversarial learning and utilized style-wise loss for training, yet the new method also used semi-supervised loss for adversarial learning. An advantage of the new model over the ARPM-net is achieved by synthesizing additional training data and leveraging the adversarial loss from the trustworthy area on the confidence map.

The new method can be potentially improved. Firstly, like most adversarial learning models, its hyperparameters are delicate to tune. Training the model may encounter mode collapse, and it is challenging to balance the learning progress of the segmentation network and the discriminator. Secondly, the potential for progressive growth GAN [18] has not been fully exploited. For instance, it is possible to generate new data with corresponding ground-truth annotations [25]. Generating accurately annotated medical images remains a hard problem, so we plan to investigate how synthesized annotated CT images can aid the learning process.

## 6. Conclusion

The research was motivated in part to address the problem of the lack of sufficient data in medical imaging. We proposed and developed a semi-supervised adversarial learning framework supported by data augmentation using GAN for 3D male pelvic CT segmentation, which is also expected to boost a wide range of medical machine learning applications currently suffer the lack-of-data bottleneck. We cast semantic segmentation as a problem of adversarial learning for data generation and semi-supervised learning to use both annotated and un-annotated data. The results, quantified by four popular segmentation quality metrics, showed that our new method achieved state-of-the-art performance evaluated on a set of clinical 3D pelvic CT images and 10-fold-cross validation.

A comparative analysis showed that the new semi-supervised adversarial learning method outperformed the currently best methods with the same amount of annotated data. In particular, the new method achieved comparable performance with less annotated data. It can produce high-quality contours for five organs (the prostate, bladder, rectum, left femur, and right femur).

**Acknowledgment**

The work was supported in part by Varian Medical Systems through a research grant.